\documentclass[letterpaper, 10 pt, conference]{ieeeconf}  
\pdfoutput=1

\IEEEoverridecommandlockouts                              

\usepackage{cite}
\usepackage{color}
\usepackage{latexsym}
\usepackage{amsmath}
\usepackage{amssymb}
\usepackage{algorithm}
\usepackage{algorithmic}
\usepackage{graphics} 
\usepackage{epsfig} 
\usepackage{booktabs} 
\usepackage[colorlinks,linkcolor=blue]{hyperref}

\renewcommand{\b}{\mathbf{b}}

\newcommand{\g}{\mathbf{g}}

\renewcommand{\u}{\mathbf{u}}

\renewcommand{\v}{\mathbf{v}}

\newcommand{\w}{\mathbf{w}}

\newcommand{\norm}[1]{\mbox{$\left\lVert #1 \right\rVert$}}

\newtheorem{thm}{Theorem}

\newtheorem{lem}{Lemma}

\newtheorem{prop}[thm]{Proposition}

\newtheorem{defn}{Definition}

\newtheorem{rem}{Remark}

\newtheorem{assum}{Assumption}

\title{\LARGE \bf
Byzantine-Resilient Stochastic Gradient Descent for Distributed Learning: A Lipschitz-Inspired Coordinate-wise Median Approach
}

\author{Haibo Yang, Xin Zhang, Minghong Fang, and Jia Liu
\thanks{Haibo Yang, Minghong Fang, and Jia Liu are with the Department of Computer Science, Iowa State University, Ames, IA 50011, USA
        {\tt\small yanghb@iastate.edu, myfang@iastate.edu, jialiu@iastate.edu}}%
\thanks{Xin Zhang is with the Department of Statistics, Iowa State University,
        Ames, IA 50011, USA
        {\tt\small xinzhang@iastate.edu}}%
}

\begin{document}


\maketitle
\thispagestyle{empty}
\pagestyle{empty}


\begin{abstract}
In this work, we consider the resilience of distributed algorithms based on stochastic gradient descent (SGD) in distributed learning with potentially Byzantine attackers, who could send arbitrary information to the parameter server to disrupt the training process.
Toward this end, we propose a new Lipschitz-inspired coordinate-wise median approach (LICM-SGD) to mitigate Byzantine attacks.
We show that our LICM-SGD algorithm can resist up to half of the workers being Byzantine attackers, while still converging almost surely to a stationary region in non-convex settings.  
Also, our LICM-SGD method does not require any information about the number of attackers and the Lipschitz constant, which makes it attractive for practical implementations. 
Moreover, our LICM-SGD method enjoys the optimal $O(md)$ computational time-complexity in the sense that the time-complexity is the same as that of the standard SGD under no attacks.  
We conduct extensive experiments to show that our LICM-SGD algorithm consistently outperforms existing methods in training multi-class logistic regression and convolutional neural networks with MNIST and CIFAR-10 datasets.
In our experiments, LICM-SGD also achieves a much faster running time thanks to its low computational time-complexity. 
\end{abstract}



\section{Introduction} \label{sec:intro}

Fueled by the rise of machine learning and big data analytics, recent years have witnessed an ever-increasing interest in solving large-scale empirical risk minimization problems (ERM) -- a fundamental optimization problem that underpins a wide range of machine learning applications.
In the post-Moore's-Law era, however, to sustain the rapidly growing computational power needs for solving large-scale ERM, the only viable solution is to exploit {\em parallelism} at and across different spatial scales.
Indeed, the recent success of machine learning applications is due in large part to the use of distributed machine learning frameworks (e.g., TensorFlow\cite{TensorFlow} and others)
which exploit the abundance of distributed CPU/GPU resources in large-scale computing clusters.
Furthermore, in many large-scale learning systems, data are sampled and stored at different geo-locations.
As a result, it is often infeasible to move all the data to a centralized location because of prohibitively high costs or privacy concerns.
Due to these factors, first-order stochastic gradient descent (SGD) based methods have been the workhorse algorithms in most distributed machine learning frameworks thanks to their low complexity and simplicity in distributed implementations.
Unfortunately, the proliferation of distributed learning systems also introduces many new cybersecurity challenges in the design of SGD-based distributed optimization algorithms.
Besides the conventional computation/communication errors or stalled processes seen in traditional distributed computing systems, a serious problem in SGD-based distributed optimization methods is that they are prone to the so-called Byzantine attacks, where a malicious worker machine returns arbitrary information to the parameter server.
It has been shown in \cite{blanchard2017machine} that, under Byzantine attacks, even a single erroneous gradient can fail the whole learning system and causing the classical distributed SGD algorithm to diverge. 
In light of the vulnerability of the traditional SGD-based optimization algorithms, there have been strong interests in designing robust SGD-type algorithms that are resilient to Byzantine attacks in distributed learning.
However, developing Byzantine-resilient optimization algorithms for distributed learning is highly non-trivial. 
Despite a significant amount of efforts spent over the years, most existing work in the literature on Byzantine-resilient distributed algorithms (see, e.g., \cite{blanchard2017machine,yin2018byzantine,mhamdi2018hidden,chen2017distributed,xie2018generalized,xie2018zeno,xie2018phocas}) have two main limitations:
i) requiring the knowledge of the number of malicious workers, which is often infeasible in practice;
and ii) suffering from high computational complexity in stochastic subgradient screening and aggregation mechanisms (see Section~\ref{sec:related_work} for detailed discussions).
Moreover, the classification accuracy performance of these existing works are far from satisfactory in practice (see our numerical experiments in Section~\ref{sec:numerical}).
The limitations of these existing works motivate us to develop a new Byzantine-resilient SGD-based optimization algorithm for distributed learning.

The main contribution of this paper is that we propose a new Byzantine-resilient SGD-based optimization algorithm based on a low-complexity Lipschitz-inspired coordinate-wise median approach (LICM-SGD), which overcomes the aforementioned limitations in mitigating Byzantine attacks.
Our main technical results are summarized as follows:
\begin{list}{\labelitemi}{\leftmargin=1em \itemindent=-0.4em \itemsep=.2em}
\item Inspired by the rationale that ``benign workers should generate stochastic gradients closely following the Lipschitz characteristics of the true gradients,'' we develop a Lipschitz-inspired coordinate-wise stochastic gradient screening and aggregation mechanism.
We show that our proposed LICM-SGD can resist up to one-half of the workers being Byzantine, and yet still achieve the same convergence performance compared to the no-attack scenario.
We note that these nice performance gains under LICM-SGD are achieved {\em without} requiring any knowledge of the number of Byzantine attackers, which is assumed in most existing works (see, e.g., \cite{blanchard2017machine,yin2018byzantine,mhamdi2018hidden,xie2018generalized,xie2018zeno,xie2018phocas,damaskinos2018asynchronous}).
Hence, our proposed LICM-SGD is more advantageous for practical implementations.
\item Another salient feature of our proposed LICM-SGD approach is that it has low computational complexity in stochastic gradient screening and aggregation.
Specifically, our LICM-SGD method only requires $O(md)$ time-complexity, where $m$ is the number of worker machines and $d$ is the dimensionality of the ERM problem. 
We note that this time-complexity is {\em optimal} in the sense that it is the same as the most basic distributed SGD algorithm, which has the lowest computational time-complexity.
In contrast, most of the existing algorithms with Byzantine-resilience performance similar to ours (i.e., being able to resist up to one half of workers being Byzantine) have time complexity $O(m^{2}d)$, which is problematic in data centers where the number of workers is typically large.

\item Last but not least, to verify the real-world performance of our proposed LICM-SGD approach, we conduct extensive experiments by training multi-class regression (MLR) and convolutional neural networks (CNN) based on the MNIST and CIFAR-10 datasets.
Our experimental results show that the classification accuracy under LICM-SGD is only $5\%$ lower than the standard distributed SGD method with no attacks, and consistently outperforms the state of the art in the literature.
For MLR and CNN training on the MNIST datasets, LICM-SGD can reach up to $85\%$ classification accuracy, which is three times as high as other coordinate-wise median-based methods.
These good numerical results corroborate our theoretical results.
\end{list}

Collectively, our results advance the design of Byzantine-resilient SGD-type optimization algorithms for distributed machine learning.
The remainder of this paper is organized as follows:
In Section~\ref{sec:related_work}, we review the literature to put our work in comparative perspectives.
Section~\ref{sec:model} introduces the system and problem formulation.
Section~\ref{sec:alg} focuses on the LICM-SGD algorithmic design and performance analysis.
Section~\ref{sec:numerical} presents numerical results and Section~\ref{sec:conclusion} concludes this paper.

\section{Related Work} \label{sec:related_work}

Generally speaking, the basic idea behind most Byzantine-resilient SGD methods  is to take the median of a batch of stochastic gradients, which is statistically more stable than taking the average as in the basic SGD method.
Under this basic idea, existing works can be roughly classified into two main categories as follows.

The first category is based on the geometric median. 
Specifically, this approach aims to find a point in the parameter vector space that minimizes the sum distance to the current batch of stochastic gradients in some $\ell_p$ norm sense. 
Two notable approaches in this category are \cite{chen2017distributed} and \cite{blanchard2017machine}, both of which can be shown to converge with up to half of the workers being Byzantine.
We note that these methods have the same Byzantine resilience performance compared to ours.
However, the computational complexity of these  approaches is $O(m^{2}d)$, where $m$ is the number of workers and $d$ is the dimensionality of the problem.
We note that this computational complexity is significantly higher than our $O(md)$ result and could be problematic in data center settings, where the number of machines is on the order of thousands or even higher.
Further, it has been recently shown that Byzantine attackers can utilize the high dimensionality and the highly non-convex landscape of the loss function in ERM to make geometric-median-based methods ineffective\cite{mhamdi2018hidden}. 
The reason is that since geometric median minimizes the sum distance of all dimensions, it may not be able to discriminate how much two stochastic gradients disagree in each dimension. 
To address this problem, the authors of \cite{mhamdi2018hidden} proposed a hybrid strategy: the first step is to recursively use a geometric-median-based method to pick $m - 2q$ gradients, and then utilize a coordinate-wise trimmed mean approach (to be discussed shortly) to determine the update direction. 
However, their computational time-complexity remains $O(m^2d)$ and the maximum number of Byzantine attackers cannot exceed $1/4$ of total workers.

The second category is based on using the coordinate-wise median (or related methods) as an aggregation rule for the stochastic gradient updates. 
Specifically, coordinate-wise median methods simply take the median in each dimension rather than the entire vector as in geometric median. 
For example, in \cite{yin2018byzantine}, the authors gave a sharp analysis on the statistical error rates for two Byzantine-resilient algorithms, namely coordinate-wise median and trimmed mean, respectively.
The convergence of these existing coordinate median based methods can tolerate up to half of the workers being Byzantine, and with computational time-complexity $O(md)$ (same as ours).
However, in practice, the classification accuracy performance of these coordinate-wise median based methods are usually worse than those  of the geometric median based methods.
We note that our LICM-SGD method also falls into the category of coordinate-wise median methods.
However, our work differs from the existing coordinate median approaches in that we develop a Lipschitz-inspired selection rule for screening stochastic gradients.
Our LICM-SGD method not only retains the low $O(md)$ time-complexity, but also achieves a classification accuracy significantly higher than the existing work.

We note that there are other related works that consider different settings under Byzantine attacks, which are not directly comparable to our work algorithmically. 
For example, the authors in \cite{damaskinos2018asynchronous} considered asynchronous SGD, while the authors in\cite{alistarh2018byzantine} used historical information to remove suspicious workers from future considerations. 
In \cite{cao2018distributed}, the authors proposed to let the parameter server keep a small set of data to compute an estimate of the true gradient, which is used as a benchmark to filter out suspicious gradients. 
In \cite{xie2018zeno}, the authors developed a suspicion-based aggregation rule, but with a weaker attack model that the Byzantine attackers have no information about the loss function.


\section{System Model and Problem Statement} \label{sec:model}

\begin{figure}[!t]
\vspace{0mm}
\centering
{\includegraphics[width=0.8\columnwidth]{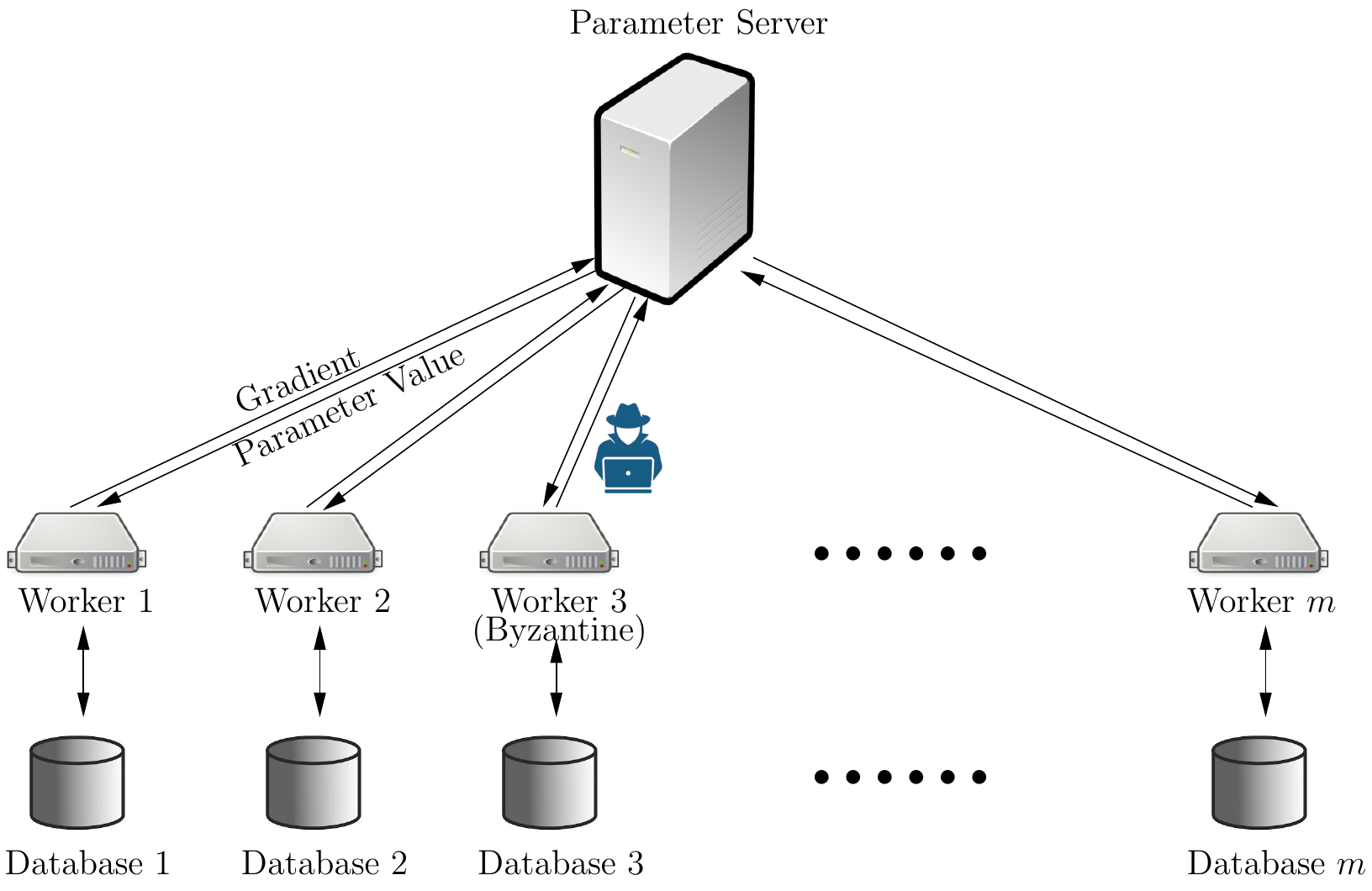}}
\caption{A distributed learning system with Byzantine attacks.}
\label{fig_SysArch}
\vspace{-.2in}
\end{figure}

{\bf Notation: } In this paper, we use boldface to denote matrices/vectors.
We let $[\v]_{i}$ represent the $i$-th entry of $\v$.
We use $\norm{\cdot}$ to denote the $\ell^{2}$-norm.

In this paper, we consider a distributed machine learning system with potential adversarial attacks.
As shown in Fig.~\ref{fig_SysArch}, there are one parameter server (PS) and $m$ distributed worker machines in the system.
Some of the workers could be malicious and launch Byzantine attacks.
We consider a standard empirical risk minimization (ERM) setting, where the goal is to find an optimal parameter vector $\w^{*}$ that minimizes the risk function  $F(\w)$, i.e.,
\begin{align*}
\w^{*} = \arg\min_{\w \in \mathbb{R}^{d}} F(\w) \triangleq \frac{1}{N} \sum_{j=1}^{N} f(\w, \xi_j),
\end{align*}
where $f: \mathbb{R}^{d} \rightarrow \mathbb{R}$ represents a loss function of some learning models (e.g., regression, deep neural networks, etc.) and $\{ \xi_{j} \}_{j=1}^{N}$ denotes the set of random samples with size $N$.
In this system, the PS and workers employ a stochastic gradient descent (SGD) based optimization algorithm to solve the ERM problem in a distributed fashion as follows:
In iteration $k$, each worker $i$ first retrieves the current parameter value $\w_{k}$ from the PS.
It then draws a mini-batch of independently and identically distributed (i.i.d.) random data samples from its local database to compute a stochastic gradient $g_{i}(\w_{k}) \in \mathbb{R}^{d}$ of the loss function and return $g_{i}(\w_{k})$ to the PS.
Upon the reception of all $m$ stochastic gradients from the workers, the PS updates the parameter vector as follows:
\begin{align} \label{eqn_sgd_generic}
\w_{k+1} = \w_k - \eta_k G[g_i(\w_k),\ldots,g_m(\w_{k})],
\end{align}
where $\eta_{k}$ is the step-size in iteration $k$ and $G[\cdot]$ denotes some aggregation rule.
As an example, for the standard SGD in normal systems without attacks, $G$ corresponds to taking the average of the stochastic gradients.
 
In Byzantine attacks, malicious workers have the full knowledge of the system and they can collaborate with each other.
Each Byzantine worker could return arbitrary values to the PS.
Hence, the Byzantine attack model is viewed as the most difficult class of attacks to defend in distributed systems.
In this paper, we aim to {\em develop a robust aggregation function $G(\cdot)$ (instead of taking the average) to mitigate Byzantine attacks}.
Toward this end, we make the following assumptions:

\smallskip
\begin{assum}[Unbiased Gradient Estimator] \label{assum1}
The stoc- hastic gradient $g_i(\w_{k})$ returned by a correct (non-Byzantine) worker $i$ is an unbiased estimator of the true gradient of $F(\cdot)$ evaluated at $\w_{k}$, i.e., $\mathbb{E} g_{i}(\w_k) = \nabla F(\w_k)$, $i \in \{1,\ldots,m\}$.
\end{assum}

\smallskip
\begin{assum}[Bounded Variance] \label{assum2}
The stochastic gradient $g_i(\w_{k})$ from a correct (non-Byzantine) worker $i\in\{1,\ldots,m\}$ has a bounded variance, i.e., $\mathbb{E} \big| [g_i(\w_k)]_{j} - [\nabla F(\w_k)]_j \big|^2 \!\leq\! \sigma ^2$, $\forall j \!=\! 1,\ldots,d$, where $\sigma^{2} \!>\! 0$ is a constant.
\end{assum}

\smallskip
\begin{assum}[Lipschitz Continuous Gradient] \label{assum3}
There exists a constant $L>0$ such that for all $\w_{1}, \w_{2} \in \mathbb{R}^{d}$, it holds that $\norm{ \nabla F(\w_1) - \nabla F(\w_2) } \leq L \norm{ \w_1 - \w_2 }$.
\end{assum}

\smallskip
\begin{assum}[Step-size] \label{assum4}
The step-sizes $\{\eta_{k}\}$ in (\ref{eqn_sgd_generic}) satisfy
$\sum_{k = 1}^{\infty} \eta_k = \infty$ and $\sum_{k = 1}^{\infty} \eta_k^2 < \infty$.
\end{assum}

\smallskip
\begin{assum}[Differentiability] \label{assum5}
The objective function $F(\cdot)$ is three times differentiable with continuous derivatives.
\end{assum}

\smallskip
\begin{assum}[Linear Growth of the $r$-th Moment] \label{assum6}
For\\ a stochastic gradient $g_i(\w_{k})$ returned by a correct (non-Byzantine) worker $i$, there exist positive constants $C_r, D_{r} > 0$, $r=2,3,4$, such that 
$\mathbb{E} \norm {g_i(\w_k) }^r \leq C_r + D_r \norm{\w_k }^r, \forall \w_k \in \mathbb{R}^d$.
\end{assum}

\smallskip
Several remarks for Assumptions~\ref{assum1}--\ref{assum6} are in order:
Assumptions~\ref{assum1}--\ref{assum4} are standard in the SGD convergence analysis literature. 
Assumption~\ref{assum5} can usually be satisfied by most learning problems in practice.
Assumption~\ref{assum6} implies that the $r$-th moment of a non-Byzantine stochastic gradient does not grow faster than linearly with respect to the norm of $\w_{k}$, which is a necessary condition in order to bound the distribution tails  for convergence (cf. \cite{bottou1998online}). 
We note that Assumption~\ref{assum6} is not restrictive and has appeared in non-convex stochastic approximations (e.g., \cite{bottou1998online}) as well as several recent Byzantine tolerant gradient descent algorithms \cite{blanchard2017machine} \cite{damaskinos2018asynchronous}. 
Note that we do not assume any convexity property about $F(\cdot)$, i.e., $F(\cdot)$ could potentially be non-convex.

\smallskip
With the above modeling and assumptions, we are now in a position to present our Byzantine-resilient SGD method, which constitutes the main subject in the next section.



\section{A Lipschitz-Inspired Coordinate-wise Median Approach to Mitigate Byzantine Attacks} \label{sec:alg}

In this section, we will first introduce our LICM-SGD algorithm in Section~\ref{subsec:licm}.
Then, we will present the main theoretical results and their intuitions in Section~\ref{subsec:results}.
The proofs for the main results are provided in Section~\ref{subsec:proofs}.

\subsection{The LICM-SGD Algorithm} \label{subsec:licm}

Before presenting our LICM-SGD algorithm, we first formally define the notion of coordinate-wise median, which serves as a key building block for our aggregation rule.

\smallskip
\begin{defn}[Coordinate-wise Median] 
For a set of vectors $\v_{i} \in \mathbb{R}^d, i = 1,\ldots,m$, the coordinate-wise median, denoted as $\mathtt{CoorMed} \lbrace \v_i, i = 1,\ldots,m \rbrace$, is a vector with its $j$-th coordinate being $\mathtt{Med} \lbrace [\v_i]_{j}, i = 1,\ldots,m \rbrace, j =1,\ldots,d$, where $\mathtt{Med}\{\cdot\}$ denotes the median of a set of scalars.
\end{defn}

\smallskip
Our proposed LICM-SGD algorithm is stated in Algorithm~1 as follows:
\bigskip
\hrule
\vspace{.03in}
\noindent {\bf Algorithm~1:} The LICM-SGD Algorithm.
\hrule
\vspace{.00in}
\vspace{0.1in}
\noindent {\bf Initialization:}
\begin{enumerate} 
\item[1.] Let $k=0$. Choose an initial parameter vector $\w_0$ and an initial step-size $\eta_0$. 
\end{enumerate}

\noindent {\bf Main Loop:}
\begin{enumerate} 
\item[2.] In the $k$-th iteration, each worker $i$ retrieves the current parameter vector $\w_k$ from the PS, $\forall i=1,\ldots,m$.

\item[3.] For each worker $i\in\{1,\ldots,m\}$, if it is non-Byzantine, it computes a stochastic gradient $g_{i}(\w_k)$ based on a mini-batch and returns $g_{i}(\w_{k})$ to the PS; otherwise, it returns an arbitrary value to the PS.

\item[4.] Upon receiving stochastic gradients from all $m$ workers, the PS computes the coordinate-wise median $\u_{k}$:
\begin{align} \label{eqn_CoordMed}
\u_{k} = \mathtt{CoorMed} \{ g_i(\w_k), i=1,\ldots,m \}.
\end{align}
Then, the PS selects all stochastic gradient vectors $g_{i}(\w_{k})$, $i \in \{1,\ldots,m\}$, that satisfy: 
\begin{align} \label{eqn_selection_rule}
\big| [g_{i}(\w_k)]_{j} \!-\! [\u_{k-1}]_{j} \big| \!\leq\! \gamma \big| [\u_{k}]_{j} \!-\! [\u_{k-1}]_{j} \big|, 
\end{align}
$\forall j \!\in\! \{1,\ldots,d\}$,
to form a set $\mathcal{T}_{k}^{*}$,
where $\gamma \geq 1$ is some system parameter.
If $k=0$, let $\tilde{g}(\w_0) = \u_{0}$. Otherwise, let 
\begin{align} \label{eqn_aggregation}
\tilde{g}(\w_k) = \frac{1}{| \mathcal{T}_{k}^{*} |} \sum_{i \in \mathcal{T}_{k}^{*}} g_{i}(\w_{k}).
\end{align}

\item[5.] Update the parameter vector as:
\begin{align} \label{eqn_weight_update}
\w_{k+1} = \w_{k} - \eta_{k} \tilde{g} (\w_{k}).
\end{align}

\item[6.] 
Let $k \leftarrow k+1$ and go to Step 2. 
\end{enumerate}
\smallskip
\hrule
\medskip

\smallskip
\begin{rem}
Several important remarks on Algorithm~1 are in order:
i) Algorithm~1 is inspired by the intuition that ``benign workers should generate stochastic gradients closely following the Lipschitz characteristics of the true gradients.''
To see this, note that if the $\mathtt{CoorMed}\{\cdot\}$ operation removes the outliers and so $\u_{k}$ is ``close'' to the average (i.e., an unbiased estimator of $\nabla F(\cdot)$), then we approximately have:
\begin{align*}
\frac{\big| [\u_k]_j - [\u_{k-1}]_j \big|}{\big| [\w_k]_j - [\w_{k-1}]_j \big|} \approx \frac{\big| [\nabla F(\w_{k})]_{j} - [\nabla F(\w_{k-1})]_{j} \big|}{\big| [\w_k]_j - [\w_{k-1}]_j \big|} \leq L,
\end{align*}
where the last inequality follows from the Lipschitz gradient assumption (cf. Assumption~\ref{assum3}).
Hence, if a stochastic gradient $g_{i}(\w_k)$ satisfies the following relationship:
\begin{multline} \label{eqn_LICM_inspiration}
\frac{| [ g_{i}(\w_{k})]_j - [\u_{k-1}]_j |}{| [\w_k]_j - [\w_{k-1}]_j |} \leq \frac{\gamma |[\u_k]_j - [\u_{k-1}]_j |}{| [\w_k]_j - [\w_{k-1}]_j |} \\
\approx \frac{ \gamma | [\nabla F(\w_{k})]_{j} - [\nabla F(\w_{k-1})]_{j} | }{| [\w_k]_j - [\w_{k-1}]_j |} \lessapprox \gamma L, \quad \forall j,
\end{multline}
for some appropriately chosen parameter $\gamma \geq 1$, then it is likely that worker $i$ is a benign worker.
Lastly, extracting the numerator relationship from (\ref{eqn_LICM_inspiration}) yields:
\begin{align*} 
\big| [g_{i}(\w_k)]_{j} - [\u_{k-1}]_{j} \big| \leq \gamma \big| [\u_{k}]_{j} - [\u_{k-1}]_{j} \big|,
\end{align*}
which is the stochastic gradient selection rule (\ref{eqn_selection_rule}) in Algorithm~1.
It is also worth pointing out that even though Eq.~(\ref{eqn_selection_rule}) is inspired by the Lipschitz characteristics of $\nabla F(\cdot)$, we do {\em not} need to know the value of the Lipschitz constant $L$.
ii) The parameter $\gamma \geq 1$ is used to balance the trade-off between Byzantine-resilience and variance.
This is because, on one hand, increasing $\gamma$ helps increase the size of $\mathcal{T}_{k}^{*}$, leading to smaller variance (if all included stochastic gradients are benign) but at a higher risk to include Byzantine workers.
On the other hand, decreasing $\gamma$ could lead to too few stochastic gradients being eligible, which results in a larger variance in the final stochastic gradient computation (\ref{eqn_aggregation}).
In practical implementations, $\gamma$ can either be preset or data-driven (i.e., adaptively choosing $\gamma$ based on the sequentially arriving random samples).
iii) We note that, unlike most existing methods in the literature, Algorithm~1 does {\em not} require the knowledge of the number of Byzantine workers, which is usually difficult to estimate in practice.
Hence, our proposed LICM-SGD method is more advantageous for practical implementations.
\end{rem}

\subsection{Main Theoretical Results} \label{subsec:results}

For better readability, we summarize the main theoretical results in this subsection and relegate their proofs to Section~\ref{subsec:proofs}.
Our first key result in this paper suggests that the coordinate-wise median of a set of stochastic gradients under Byzantine attacks shares the {\em same} statistical characteristics as that of a non-Byzantine stochastic gradient if the number of Byzantine workers is less than half of the total.

\smallskip
\begin{lem}[Statistical Properties of $\mathtt{CoorMed}(\cdot)$] \label{lem_coordmed_stat}
Let $q$ be the number of Byzantine workers and if $2q + 1 < m$, then $\u_{k} = \mathtt{CoorMed} \{ g_i(\w_k), i = 1,\ldots,m \}$ share the same statistical characteristics as a stochastic gradient returned by a non-Byzantine worker:
\begin{align*}
&\mathbb{E} \big| [\u_{k}]_{j} -[ \nabla F(\w_k)]_j \big|^2 \leq \sigma ^2, \quad j=1,\ldots,d, \\
&\mathbb{E} \norm{\u_k}^r \leq C_r + D_r\norm{\w_k}^r, \quad \forall \w_k \in \mathbb{R}^d, r = 2,3,4.
\end{align*} 
\end{lem}

\smallskip
Lemma~\ref{lem_coordmed_stat} indicates that, statistically, $\u_k$ can be viewed as a non-Byzantine stochastic gradient if $2q + 1 < m$. 
This insight is our fundamental rationale to use $\u_k$ as a benchmark for our aggregation criterion in Algorithm~1.
Thanks to this nice statistical property of coordinate-wise median, we can further prove that, based on the screening rule in (\ref{eqn_selection_rule}), the aggregated vector $\tilde{g}(\w_{k})$ obtained from (\ref{eqn_aggregation}) has linear growths of the $r$-th moments, $r=2,3,4$.
We formally stated this result in the following lemma:
\begin{lem}[Linear Growth of $r$-th Moment] \label{lem_rth_moment_aggr} There exists constants $A_r ,B_r > 0$, $r=2,3,4$, such that $\forall k \geq 0, \mathbb{E} \norm{ \tilde{g}(\w_k) }^r \leq A_r + B_r \norm{\w_k}^r$ if the step-sizes $\eta_k$, $\forall k$, satisfy $\eta_k \leq \min \lbrace h_r(A_r, B_r,C_{r},D_{r}), r = 2, 3, 4\rbrace$, where the functions $h_r(\cdot)$, $r=2,3,4$, will be specified in the proof.
\end{lem}

\smallskip
Lemma~\ref{lem_rth_moment_aggr} indicates that, under the aggregation rule in Algorithm~1, the obtained $\tilde{g}(\w_{k})$ remains satisfying the linear growth of the $r$-th moment assumption (cf. Assumption~\ref{assum6}).
As will be discussed next, the boundedness of the moments implies the global confinement property of the weight parameter vector $\w_{k}$, which in turn will play an important role in establishing convergence of SGD-type methods in non-convex optimization.
We formally state the global confinement property as follows:

\smallskip
\begin{lem}[Global Confinement of $\{\w_{k}\}$] \label{lem_GlobalConf}
Let $\{\w_k\}$ be the sequence of weight parameter vectors generated by Algorithm~1. 
There exist a constant $D > 0$ such that the sequence $\{\w_k\}$ almost surely satisfies $\norm{\w_k} \leq D$ as $k \rightarrow \infty$.
\end{lem}

\smallskip
In what follows, we will characterize the Byzantine-resilience performance of our proposed LICM-SGD approach.
Toward this end, we first introduce a useful performance metric for measuring Byzantine resilience:

\smallskip
\begin{defn}[($\alpha, q$)-Byzantine Resilience \cite{blanchard2017machine}] \label{defn_BR}
Let $\alpha \in [0,\pi/2]$ and let $q \in [0,m]$ be an integer. 
Let $\w_1, \ldots,\w_m$ be i.i.d. random vectors distributed as $\w \in \mathbb{R}^d$ with $\mathbb{E} \w = \g$, where $\g \in \mathbb{R}^{d}$ is deterministic. 
Let $\b_1, \b_2,\ldots,\b_q \in \mathbb{R}^d$ be any random vectors possibly dependent on $\{\w_i\}$. 
An aggregation rule $G$ is said to be ($\alpha, q$)-Byzantine resilient, if for any $1 \leq i_1 < ... < i_q \leq m$, the aggregated vector 
\begin{align*}
G(\w_1,\ldots,\underbrace{\b_1}_{i_1},\ldots, \underbrace{\b_q}_{i_q}, \ldots,\w_m)
\end{align*}
satisfies: 
i) $\langle \mathbb{E}G, \g \rangle \geq (1 - \sin \alpha)\norm{\g}^2 > 0$, 
and ii) for $r = 2,3,4$, $\mathbb{E} \norm{G}^r$ is bounded by a linear combination of terms $\mathbb{E} \norm{\w}^{r_i}$, $i=1,\ldots,n-1$, with $r_1 + ... + r_{n-1} = r$.
\end{defn}

\smallskip
Geometrically speaking, Condition i) in Definition~\ref{defn_BR} means that, with $q$ Byzantine attackers, the average of stochastic aggregation outcome of $G(\cdot)$ is contained in an error ball centered at the end point of $\g$, such that the angle between $\mathbb{E}G$ and $\g$ is bounded by $\alpha$.
Condition ii) is a technical condition that is useful in our subsequent convergence analysis.
With Lemmas~\ref{lem_coordmed_stat}--\ref{lem_rth_moment_aggr}, and Definition~\ref{defn_BR}, we are now in a position to characterize the Byzantine-resilience of our proposed LICM-SGD method as follows:

\smallskip
\begin{thm}[Byzantine Resilience] \label{thm_BR}
Under Assumptions~\ref{assum1}--\ref{assum6}, if i) the number of Byzantine worker $q$ satisfies $2q + 1 < m$ (i.e., less than one half) and ii) $(3 + 2 \gamma + \epsilon) \sqrt{d} \sigma_{k} < ||\nabla F(\w_{k})||$, $\forall k$, where $\epsilon>0$ can be arbitrarily small, then Algorithm~1 is $(\alpha ,q)-$Byzantine resilient, where $\alpha \in [0,  \pi/2]$ is defined by 
\begin{align*}
\sin \alpha = \sup_{\forall k} \bigg\{ \frac{ (3 + 2 \gamma + \epsilon) \sqrt{d} \sigma_{k}}{\norm{\nabla F(\w_k)}} \bigg\},
\end{align*}
where $\sigma_{k}$ denotes the deviation in iteration $k$.
\end{thm}

\smallskip
Based on the Byzantine resilience result in Theorem~\ref{thm_BR}, we establish the convergence of LICM-SGD as follows:

\smallskip
\begin{thm}[Almost Sure Convergence of LICM-SGD] \label{thm_as_converge}
Under Assumptions~\ref{assum1}--\ref{assum6}, the sequence of true gradients $\{ \nabla F(\w_k) \}$ generated by Algorithm~1 converges almost surely to a flat region defined by $\big\{ \w \in \mathbb{R}^{d}: \norm{\nabla F(\w)} \leq (3+2\gamma+\epsilon) \sqrt{d} \sigma_k \big\}$, where $\epsilon>0$ can be arbitrarily small.
\end{thm}

\smallskip
Finally, the time-complexity of the aggregation scheme in Algorithm~1 is stated in the following proposition:

\smallskip
\begin{prop}[Aggregation Time-Complexity] \label{prop_TimeComplexity}
The aggr- egation rule (\ref{eqn_selection_rule})--(\ref{eqn_aggregation}) in Algorithm~1 has a computational time-complexity $O(md)$.
\end{prop}

The proof, experiment setup details and the attack models are in the supplementary material. 


\section{Numerical Results} \label{sec:numerical}

In this section, we conduct experiments to show the convergence and robustness of our LICM-SGD algorithm and compare it with state-of-art Byzantine-resilient algorithms. 
We use the normal averaging method without any attacks as the ground truth baseline, denoted by ``Mean (no attack).'' 
In addition, we compare our method with state-of-art Byzantine tolerant algorithms, which can be categorized into two classes.
The first class is geometric-median-based algorithms, including Krum \cite{blanchard2017machine} and Bulyan \cite{mhamdi2018hidden}. 
The second class is coordinate-wise median-based algorithms, including Median and Trimmed mean \cite{yin2018byzantine}.

\begin{figure*}[t!]
	\begin{minipage}[t]{0.32\linewidth}
	\centering
	{\includegraphics[width=1.1\textwidth]{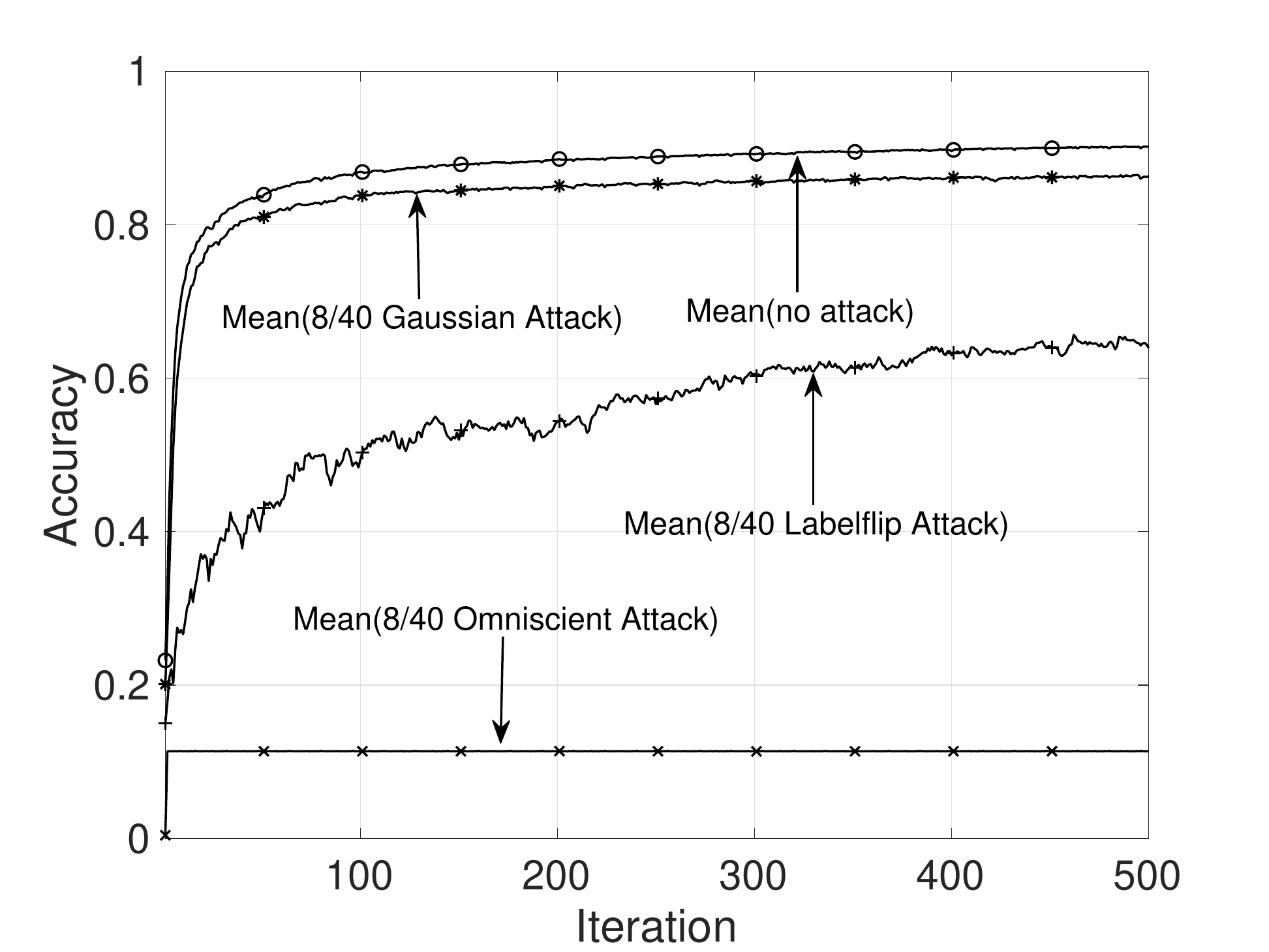}}
	\caption{MLR with different Byzantine attacks on MNIST, $q = 8, m = 40$.} 
	\label{fig1}
	\end{minipage}%
\hspace{0.009\textwidth}
	\begin{minipage}[t]{0.32\linewidth}
	\centering
	{\includegraphics[width=1.1\textwidth]{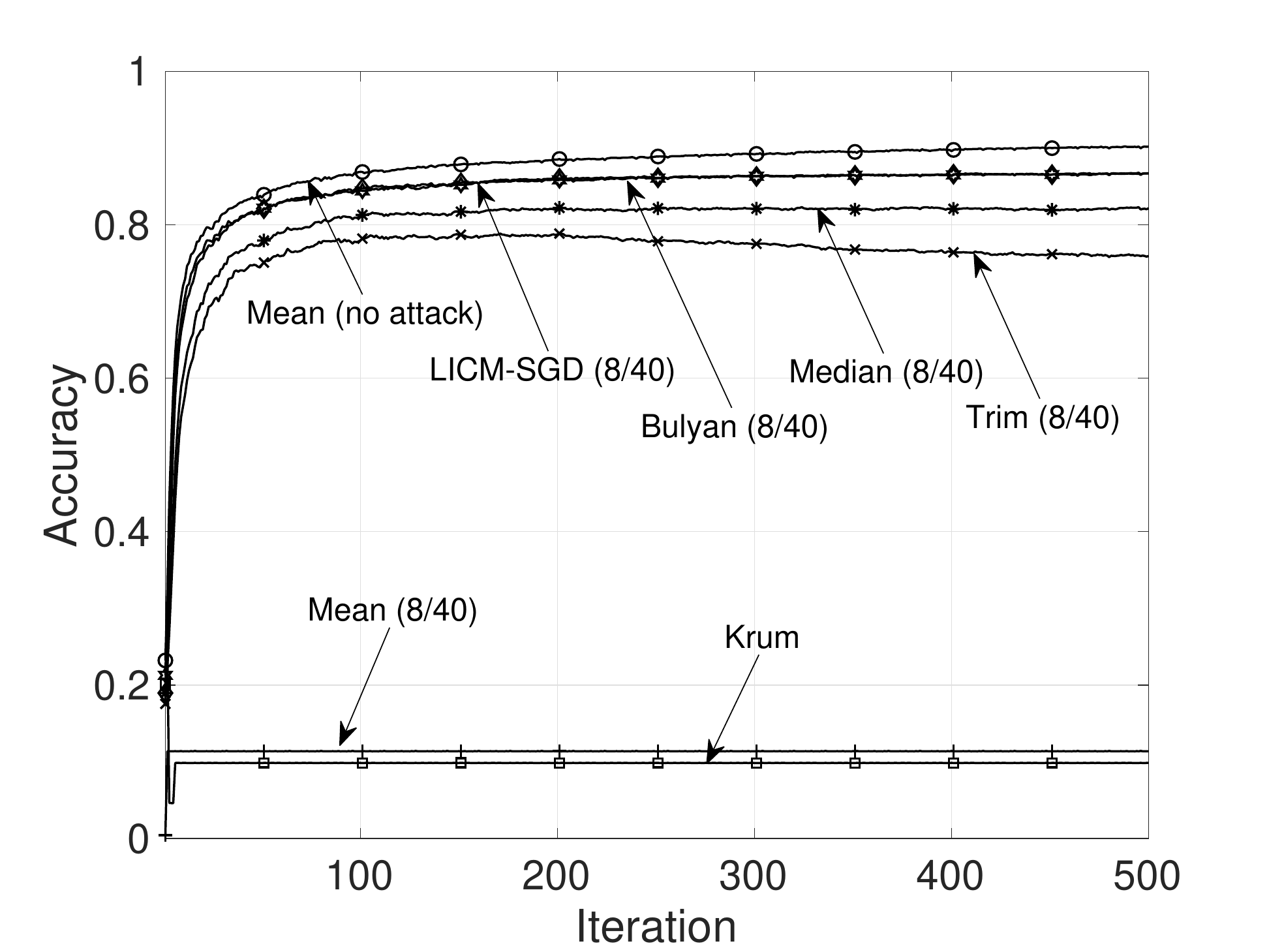}}
	\caption{MLR with Omniscient Attacks on MNIST, $q = 8, m = 40$.} 
	\label{fig2}
	\end{minipage}
\hspace{0.005\linewidth}
	\begin{minipage}[t]{0.32\linewidth}
	\centering
	{\includegraphics[width=1.1\textwidth]{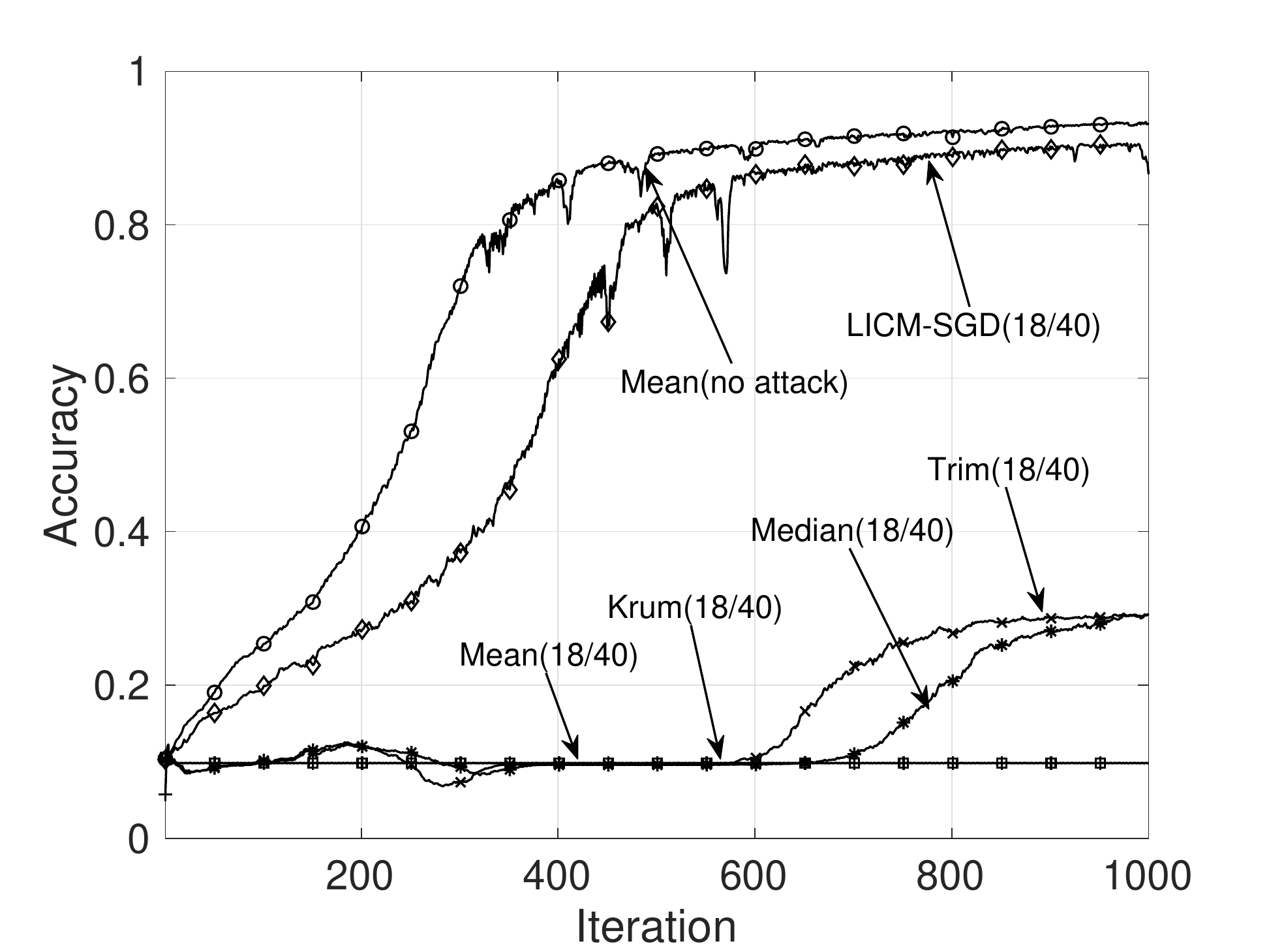}}
	\caption{CNN with Omniscient Attacks on MNIST, $q = 18, m = 40$.} 
	\label{fig3}
	\end{minipage}
\vspace{-.18in}
\end{figure*}

In Fig.~\ref{fig1}, we evaluate the effects of three different attacks on Mean method and found that even the Mean method can tolerance small fraction of Gaussian and Label Flipping attacks. \footnotemark[1]
From Fig.~\ref{fig1} and prior results\cite{blanchard2017machine,xie2018generalized}, it can be seen that the Omniscient Attacks model is much stronger than the other two. 
Thus, in the rest of the experiments, we only show the results of Omniscient Attacks.

\begin{figure*}[t!]
	\begin{minipage}[t]{0.32\linewidth}
	\centering
	{\includegraphics[width=1.1\textwidth]{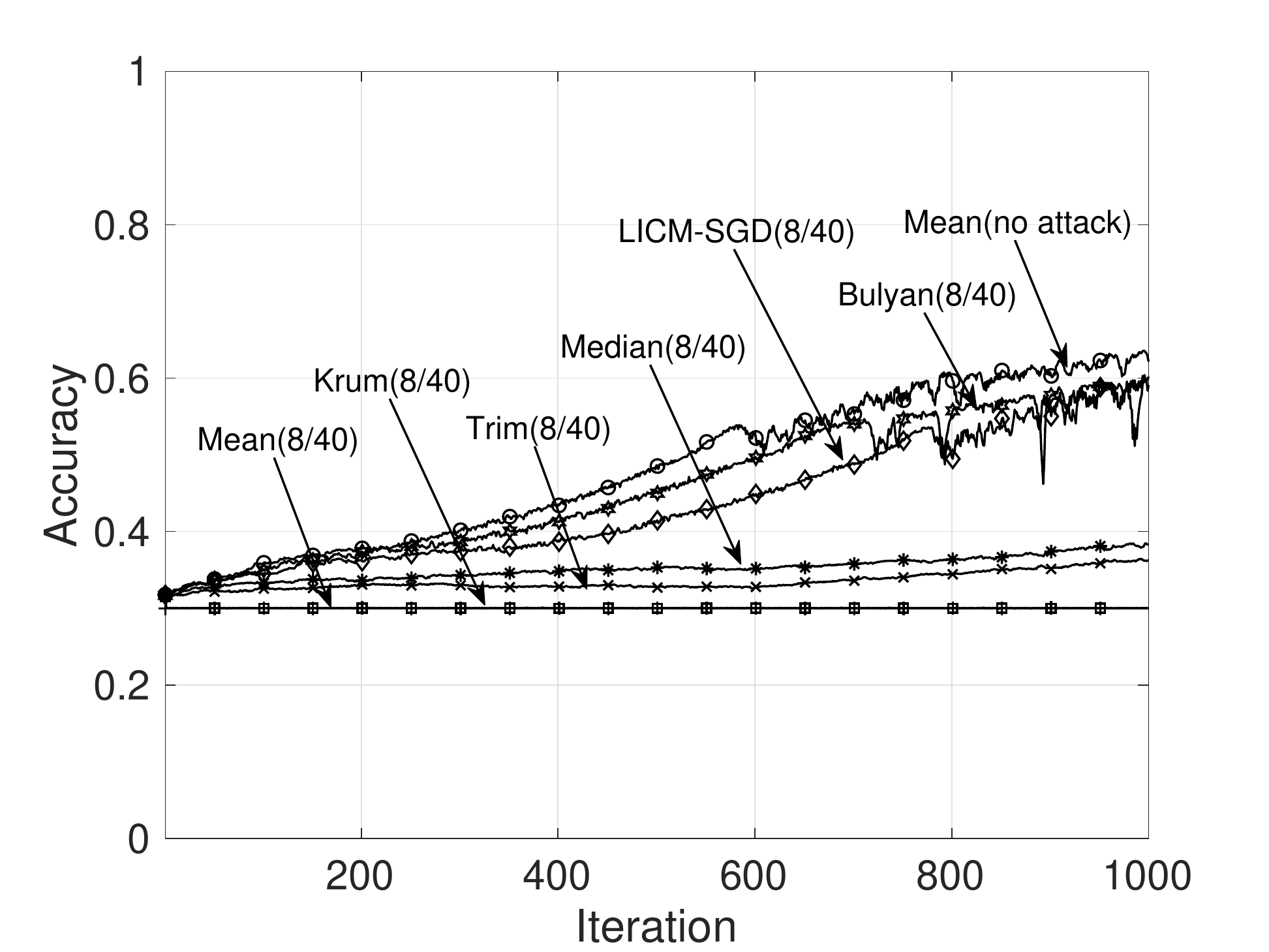}}
	\caption{CNN with Omniscient Attacks on CIFAR-10, $q = 8, m = 40$.} 
	\label{fig4}
	\end{minipage}%
\hspace{0.009\textwidth}
	\begin{minipage}[t]{0.32\linewidth}
	\centering
	{\includegraphics[width=1.1\textwidth]{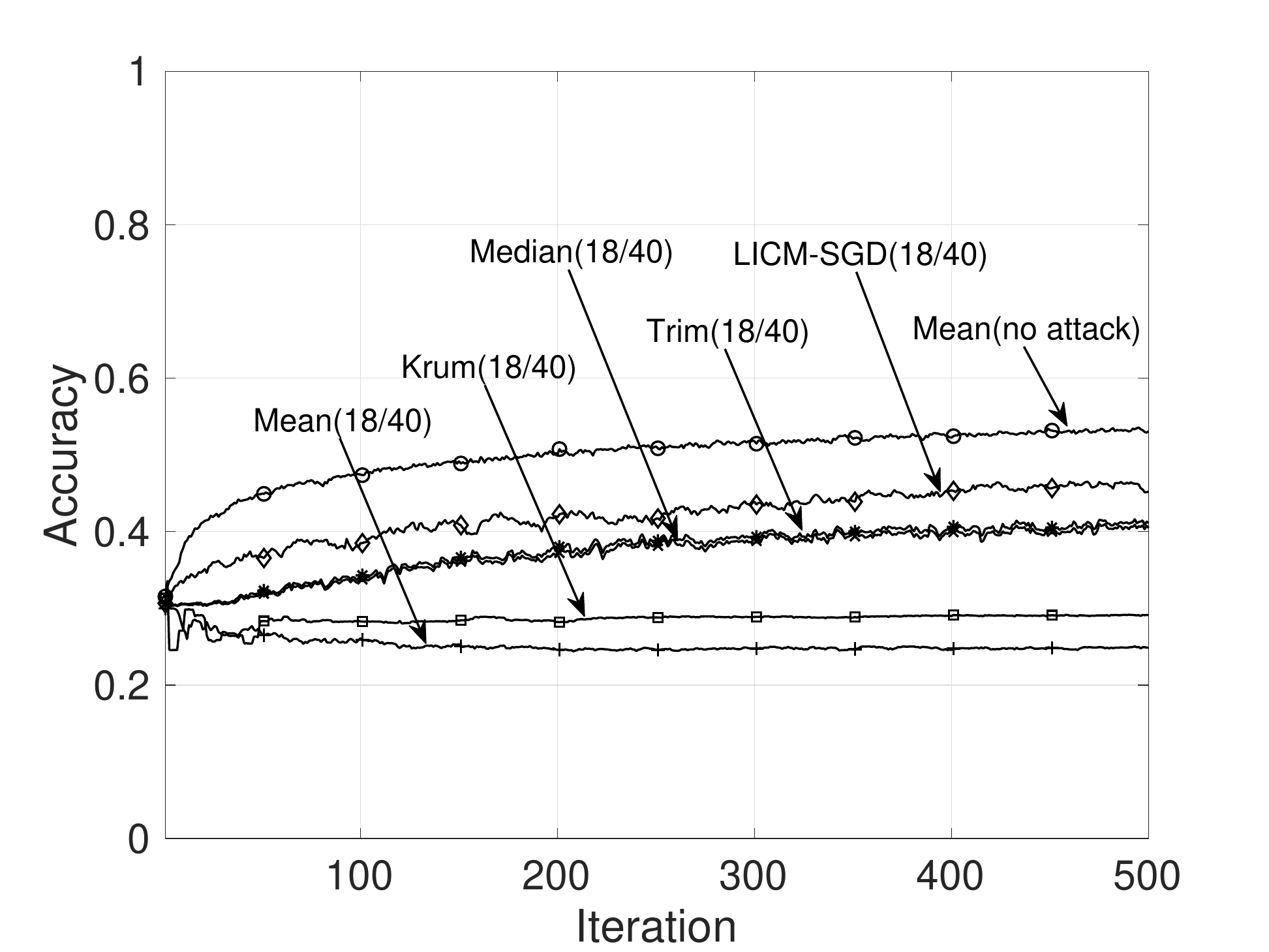}}
	\caption{MLR with Omniscient Attacks on CIFAR-10, $q = 18, m = 40$.} 
	\label{fig5}
	\end{minipage}
\hspace{0.005\textwidth}
	\begin{minipage}[t]{0.32\linewidth}
	\centering
	{\includegraphics[width=1.1\textwidth]{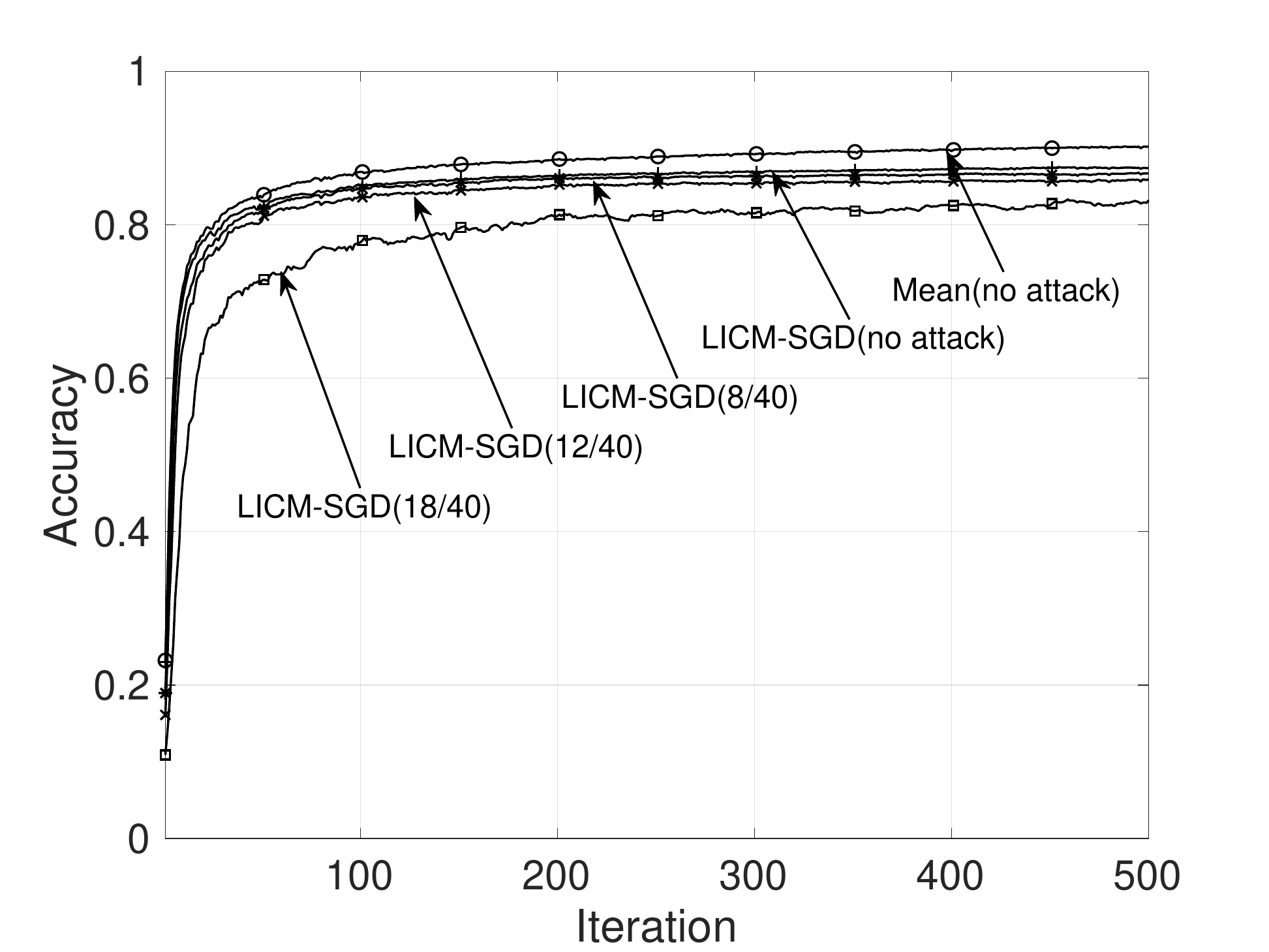}}
	\caption{MLR with Omniscient Attacks on MNIST for LICM-SGD, $q = 0,8,12,18, m = 40$.}
	\label{fig6}
	\end{minipage}
\vspace{-.2in}
\end{figure*}

\smallskip
{\bf Results:}
{\em 1) Resilience to Byzantine attacks:} 
In Fig.~\ref{fig2}, we test MLR task on MNIST with a small fraction of attackers $q = 8, m = 40$. 
Not surprisingly, all Byzantine-resilient algorithms can reach reasonable accuracy. 
Our LICM-SGD algorithm has the same accuracy as Bulyan, outperforming others. 
In Fig.~\ref{fig3}, we illustrate the CNN results for MNIST with $q = 18, m = 40$, which is the maximum defendable number of attackers $(2 q + 1 < m)$. 
All other methods fail completely with accuracy of less than $30\%$. 
Our LICM-SGD algorithm achieves $85\%$ accuracy, only slightly lower than the ground truth baseline. 
Here we do not compare with Bulyan since it requires $4q + 3 < m$, hence at most $1/4$ being Byzantine asymptotically.
Hence, Bulyan cannot be directly compared to LICM-SGD, which allows up to $1/2$  of workers being Byzantine.
We can obtain similar results on dataset CIFAR-10 for both MLR and CNN tasks under a small and large number of Byzantine attackers in Figs.~\ref{fig4} and \ref{fig5}.
We note that in all experiments, the performance of Krum\cite{blanchard2017machine} is not stable and its performance could vary significantly with different data samples.
In Figs.~\ref{fig2}--\ref{fig5}, Krum performs worse than other methods.
But we do find in other examples that Krum's performance could be comparable to other methods, which is consistent with the observations in \cite{mhamdi2018hidden,xie2018zeno}.
In contrast, the performance of LICM-SGD is insensitive to the changes in data samples.

{\em 2) Cost of Byzantine attacks:} Next, we evaluate the sensitivity of our LICM-SGD to the number of Byzantine attackers. 
Fig.~\ref{fig6} illustrates the results of MLR with Omniscient Attacks on MNIST. We can see that the presence of Byzantine attacks only has a small impact on accuracy. 
The accuracy drops from $87.5\%$ to $83.2\%$ with the attackers' number ascendant from 0 to 18, which shows that our LICM-SGD algorithm is insensitive to the number of Byzantine attackers. 
Also, LICM-SGD is a more practical algorithm than other methods since it does not require any prior information about the number of attackers. 
The number of attackers can vary greatly in practice, which could be problematic for other methods. 
Note that the value of $\gamma$ does not change in the experiment of each dataset, no matter how $q$ changes. 
In other words, $\gamma$ depends only on the dataset, but not the number of attackers. 
Note that these performances are achieved with an optimal time-complexity $O(md)$, i.e., same as the baseline mean method with no attacks.

\section{Conclusion} \label{sec:conclusion}

In this paper, we proposed a new Lipschitz-inspired coordinate-wise stochastic gradient descent method (LICM-SGD).
Our LICM-SGD method can resist up to one half of the workers being Byzantine, while still achieving the same convergence performance compared to the no-attack scenario.
Note that our LICM-SGD algorithm does not require knowledge of the number of Byzantine workers.
Thus, it is more practical compared to existing works in the literature.
Also, our proposed LICM-SGD approach enjoys a low $O(md)$ computational time-complexity, which is the same as the basic distributed SGD algorithm and significantly lower than the $O(m^2d)$ time-complexity of existing methods.
We have also conducted extensive numerical studies to verify the performance of our LICM-SGD method.
Our numerical results showed that the classification accuracy under LICM-SGD is near that of the standard distributed SGD method with no attacks (at most $5\%$ lower), and consistently performs better than existing works.
Our results in this work contribute to the increasingly important field of Byzantine-resilient distributed learning.

\bibliographystyle{IEEEtran}
\bibliography{./BIB/ByzantineReference,./BIB/DMLF,./BIB/Complete}
\section{Supplementary Material} \label{sec:proof}

\subsection{Experimental Setup}
\smallskip
{\bf Experimental Setup:}
We consider training multi-class logistic regression (MLR) and convolutional neural network (CNN) classifiers based on the MNIST and CIFAR-10 datasets.
We trained a simple convolutinoal nerral network with two convolutional layers (size $3 \times 3 \times 16$), each followed by max-pooling layer with $2 \times 2$, and a fully connected layer. ReLu activation was used for all layers and the output layer has 10 units with softmax activation.
The batch size is 32 and 64 for MLR and CNN classifiers on MNIST, respectively.
The batch size is 128 for both MLR and CNN classifiers on CIFAR-10. $\gamma$ is set to 10 in all classify tasks. In each experiment, 40 workers are launched to form the distributed learning process. 
The number of Byzantine attackers are set to 0, 8, 12, and 18, respectively.

\smallskip
{\bf Byzantine Attack Models:}
We consider three popular Byzantine attack models in this paper, namely, Gaussian Attacks, Label Flipping Attacks, and Omniscient Attacks.

\smallskip
\noindent {\em 1) Gaussian Attacks:} Byzantine attackers draw fake gradients from a Gaussian distribution with mean zero and isotropic covariance matrix with standard deviation 200 to replace the true gradient estimations.

\smallskip
\noindent {\em 2) Label Flipping Attacks:} Byzantine attackers do not require any knowledge of the training data distribution. 
In our experiment, we generate label flipping attacks as follows: On each Byzantine machine, we replace every training label $l$ with $9-l$, e.g, 2 is replaced with 7, etc.

\smallskip
\noindent {\em 3) Omniscient Attacks:} Byzantine attackers know all the benign gradients of the total $m$ workers. Then, they replace the benign gradients with the opposite value multiplied by an arbitrarily large factor.

\subsection{Proofs of the Main Results} \label{subsec:proofs}

In this subsection, we provide detailed proofs for all theoretical results stated in Section~\ref{subsec:results}.

\smallskip
{\em Proof of Lemma~\ref{lem_coordmed_stat}.}
If the number of Byzantine workers $q$ satisfies $2q+1<m$, then there must at least exist two non-Byzantine workers $p$ and $q$, such that 
\begin{align*}
[g_{p}(\w_k)]_{j} \leq [\u_{k}]_{j} \leq [g_{q}(\w_k)]_{j}.
\end{align*}
Since $p$ and $q$ are non-Byzantine, we have that $g_{p}(\w_k)$ and $g_{q}(\w_k)$ satisfy the assumptions of bounded variance and linear growth of $r$-th moment (cf. Assumption~\ref{assum2} and \ref{assum6}).
Hence, it follows that $\mathbb{E} \norm{ [\u_{k}]_{j} - [\nabla F(\w_k)]_j }^2 \leq \sigma ^2$, and $\mathbb{E} \norm{\u_k}^r \leq C_r + D_r \norm{\w_k}^r, \forall \w_k \in \mathbb{R}^d, r = 2,3,4$.
This completes the proof of Lemma~\ref{lem_coordmed_stat}.
\vspace{-.26in}
\begin{flushright}
\QEDopen
\end{flushright}

{\em Proof of Lemma~\ref{lem_rth_moment_aggr}.}
We prove Lemma~\ref{lem_rth_moment_aggr} by induction.
Let $A_r$ and $B_{r}$, $r=2,3,4$, be some constants such that $A_{r}> C_{r}$ and $B_{r}>D_{r}$, respectively.
In the base case, $\tilde{g}(\w_0) = \u_0$, which is the coordinate-wise median of all stochastic gradients in the first iteration. 
As long as the Byzantine attackers are less than half of the total workers, Lemma~\ref{lem_coordmed_stat} guarantees that $\mathbb{E} \norm{\tilde{g}(\w_0)}^r \leq  C_r + D_r \norm{\w_0}^r \leq A_r + B_r\norm{\w_0}^r, r = 2, 3, 4$.

Now, assume that in iteration $k$, we have $\mathbb{E}\norm{\tilde{g}(\w_k)}^r \leq A_r + B_r\norm{\w_k}^r$ holding true, we want to show that $\mathbb{E} \norm{ \tilde{g}(\w_{k+1})}^r \leq A_{r} + B_{r} \norm{\w_{k+1}}^r$ continues to hold in iteration $k+1$.
Toward this end, note that
\begin{align*}
&\norm{ \tilde{g}(\w_{k+1})} = \norm{ \frac{1}{|\mathcal{T}_{k}^*|} \sum_{i \in \mathcal{T}_{k}^*} g_i(\w_{k+1})} \\
&\overset{(a)}{\leq} \norm{ \frac{1}{|\mathcal{T}_{k}^*|} \sum_{i \in \mathcal{T}_{k}^*} g_i(\w_{k+1}) \!-\! \u_k} \!+\! \norm{ \u_k \!-\! \u_{k+1}} \!+\! \norm{\u_{k+1}} \\
&\overset{(b)}{\leq} (1 + \gamma) \norm{ \u_{k+1} - \u_k} + \norm{\u_{k+1}}
\end{align*}
where $(a)$ follows from the triangle inequality and $(b)$ is due to the selection process in Step 4 in Algorithm~1.
It then follows that, for $r = 2, 3, 4$, we have:
\begin{align} \label{eqn_r1}
&\norm{\tilde{g}(\w_{k+1})}^r \leq \hat{A}_r \norm{\u_{k+1} - \u_k}^r + \hat{B}_r \norm{\u_{k+1}}^r,
\end{align}
where $\hat{A}_{r}$ and $\hat{B}_{r}$ are constants that depend on $r$. 
For example, if $r = 2$, then after expanding (\ref{eqn_r1}) and collecting terms, we have $\hat{A}_{r} = 2(1+\gamma)^2$ and $\hat{B}_{r} = 2$.
The derivations for $r=3,4$ follow similar processes but are more tedious, and so we omit the details here for brevity.
From (\ref{eqn_r1}), we further have that:
\begin{align} \label{eqn_r2}
&\mathbb{E} \norm{\tilde{g}(\w_{k+1})}^r \leq \hat{A}_r \underbrace{ \mathbb{E}\norm{\u_{k+1} - \u_k}^r}_{\mathrm{(P1)}} + \hat{B}_r \underbrace{\mathbb{E}\norm{\u_{k+1}}^r}_{\mathrm{(P2)}},
\end{align}
For $r\!=\!2$, to bound (P1) in \eqref{eqn_r2}, we have from Lemma~\ref{lem_coordmed_stat} that:
\begin{align} \label{eqn_wk_wk1}
&\mathbb{E} \norm{ \u_{k+1} - \u_k }^2 \leq 2 \mathbb{E} \norm{ \u_{k+1} }^2 + 2\mathbb{E} \norm{ \u_k}^2 \nonumber \\
&\quad \leq 2(C_2 + D_2\norm{ \w_{k+1}}^2) + 2(C_2 + D_2\norm{ \w_{k}}^2) \nonumber \\
&\quad = 4C_2+ 2D_2\norm{\w_{k+1}}^2 + 2D_2\norm{\w_{k}}^2. 
\end{align}
Recall that $\w_{k+1} = \w_{k} - \eta_k \tilde{g}(\w_k)$, which implies that $\norm{\w_k} \leq \norm{\w_{k+1}} + \eta_k \norm{\tilde{g}(\w_k)}$.
It then follows that:
\begin{align} \label{eqn_wk_wk2}
\norm{\w_k}^2 &\leq 2\norm{\w_{k+1}}^2 + 2 \eta_k^2 \norm{\tilde{g}(\w_k)}^2 \nonumber\\
&\leq 2\norm{\w_{k+1}}^2 + 2 \eta_k^2 (A_2 + B_2 \norm{\w_k}^2).
\end{align}
After rearranging (\ref{eqn_wk_wk2}), we have
\begin{align*}
  (1-2B_2\eta_k^2)\norm{\w_k}^2 \leq 2\norm{\w_{k+1}}^2 + 2 A_2 \eta_k^2
\end{align*}
We assume $1 - 2B_2 \eta_k^2 > 0$, i.e. $\eta_k < \frac{1}{2B_2}$.
For notational convenience, we let $R_{1}^{(2)} \triangleq \frac{1}{2B_2}$.
Then, we have: 
\begin{align} \label{eqn_wk_bnd}
\norm{ \w_k}^2 \leq \frac{2}{ (1-2B_2\eta_k^2)}\norm{\w_{k+1}}^2 + \frac{2A_2 \eta_k^2}{(1-2B_2 \eta_k^2)}.
\end{align}
Plugging (\ref{eqn_wk_bnd}) into (\ref{eqn_wk_wk1}) yields:
\begin{align} \label{eqn_u-u_bnd}
&\mathbb{E} \norm{ \u_{k+1} - \u_k}^2 \leq 4C_2+ 2D_2\norm{ \w_{k+1}}^2 + 2D_2\norm{\w_{k}}^2 \nonumber\\
&= \Big(4C_2 \!+\! \frac{4 A_2 D_2 \eta_k^2}{1 \!-\! 2B_2 \eta_k^2}\Big) \!+\! \Big(2D_2 \!+\! \frac{4D_2}{1\!-\!2B_2\eta_t^2} \Big)\norm{\w_{k+1}}^2\!\!. \!\!\!\!\!
\end{align}
Using Lemma~\ref{lem_coordmed_stat}, we have:
\begin{align} \label{eqn_u_bnd}
\mathbb{E}\norm{\u_{k+1}}^2 &\leq C_2 + D_2 \norm{\w_{k+1}}^2 
\end{align}
Combining (\ref{eqn_u-u_bnd}) and (\ref{eqn_u_bnd}), we obtain: 
\begin{align*}
\mathbb{E} \norm{\tilde{g}(\w_{k+1})}^2 \leq A'_{2} + B'_{2} \norm{\w_{k+1}}^2,
\end{align*} 
where constants $A'_{2}$ and $B'_{2}$ are defined as:
\begin{align*}
  A'_{2} &\triangleq 2(1+\gamma)^2(4C_2 + \frac{4A_2D_2\eta_k^2}{1-2B_2\eta_k^2})+2 C_2, \\
  B'_{2} &\triangleq 2(1+\gamma)^2(4D_2 + \frac{4D_2}{1-2B_2\eta_k^2})+2 D_2.
\end{align*}
To guarantee $A'_2 \leq A_2$ and $B'_2 \leq B_2$, it suffices to have:
\begin{align*}
  \eta_k^2&\leq \frac{4A_2 -[8(1+\gamma)^2+2]C_2}{4A_2D_2 - 2B_2C_2[8(1+\gamma)^2+2]+2A_2B_2} \triangleq R_{2}^{(2)}\\
  \eta_k^2&\leq \frac{B_2 -2[6(1+\gamma)^2-1]D_2}{2B_2^2 - 8B_2D_2[(1+\gamma)^2]-4B_2D_2} \triangleq R_{3}^{(2)}
\end{align*}
Letting $h_2(A_2, B_2, C_2, D_2) \! \triangleq \! [\min(R_1^{(2)}, R_2^{(2)}, R_{3}^{(2)})]^{\frac{1}{2}}$ completes the proof for $r=2$.
For $r=3,4$, the proofs follow similar processes and we omit them in here for brevity.
\vspace{-.26in}
\begin{flushright}
\QEDopen
\end{flushright}

{\em Proof of Lemma~\ref{lem_GlobalConf}.}
With Lemma~\ref{lem_rth_moment_aggr} and Assumption~\ref{assum6}, the global confinement of $\{ \w_k \}$ follows from \cite[Sections~5.1--5.2]{bottou1998online} and \cite[Lemma~2]{damaskinos2018asynchronous}. 
\vspace{-.26in}
\begin{flushright}
\QEDopen
\end{flushright}

{\em Proof of Theorem~\ref{thm_BR}.}
To show that LICM-SGD is $(\alpha,q)$--Byzantine resilient, we first verify Condition i) in Definition~\ref{defn_BR}, i.e. $\langle \mathbb{E} \tilde{g}(\w_k), \nabla F(\w_k) \rangle \geq (1 - \sin \alpha) \norm{ \nabla F(\w_k)}^2$.
Note that
\begin{align*}
&\norm{\tilde{g}(\w_{k+1}) \!-\! \nabla F(\w_{k+1})}\overset{(a)}{\leq} \norm{ \tilde{g}(\w_{k+1}) \!-\! \u_{k}} + \norm{\u_{k+1} \!-\! \u_k} \\
&\indent + \norm{\u_{k+1} - \nabla F(\w_{k+1})} \\
& \overset{(b)}{\leq} (1 + \gamma) \norm{\u_{k+1} - \u_{k}} + \norm{\u_{k+1} - \nabla F(\w_{k+1})},
 \end{align*}
 where $(a)$ follows from the triangle inequality and $(b)$ is due to the selection process in Step 4 in Algorithm~1. 
 Following the same token, we have 
 \begin{multline*}
 \norm{\u_{k+1} - \u_k} \leq \norm{\u_{k+1} - \nabla F(\w_{k+1})} + \\
 \norm{\u_k - \nabla F(\w_k)} + \norm{ \nabla F(\w_{k+1}) - \nabla F(\w_k)}, 
\end{multline*}
which implies:
\begin{align*}
&\mathbb{E} \norm{\u_{k+1} \!-\! \u_k } \leq \mathbb{E} \norm{ \u_{k+1} \!-\! \nabla F(\w_{k+1})} \!+\! \mathbb{E} \norm{\u_{k} \!-\! \nabla F(\w_k)} \\
&+ \mathbb{E}\norm{\nabla F(\w_{k+1}) - \nabla F(\w_k)} \overset{(a)}{\leq} 2\sqrt{d} \sigma + L \eta_k \mathbb{E}\norm{\tilde{g}(\w_k)},
\end{align*}
where $(a)$ follows from Lemma~\ref{lem_coordmed_stat}, Assumption~\ref{assum2} and the update in (\ref{eqn_weight_update}).
It then follows that
\begin{multline*}
  \mathbb{E} \norm{ \tilde{g}(\w_{k+1}) - \nabla F(\w_{k+1})} \leq (3 + 2 \gamma) \sqrt{d} \sigma \\
+ (1 + \gamma) L \eta_k \mathbb{E} \tilde{g}(\w_k).
\end{multline*}
By Lemma~\ref{lem_GlobalConf}, $\w_k$ is global confined.
Hence, all continuous functions of $\w_k$ are bounded, including $\norm{\w_k}^2$, $\mathbb{E} \tilde{g}(\w_k)$, and all derivatives of the cost function $F(\w_k)$.

Since $\lim_{k \rightarrow \infty} \eta_k = 0$, for any $\epsilon > 0$, 
there exists a $k_\epsilon$, such that $\forall k \geq k_{\epsilon}$, 
$(1 + \gamma)L \eta_k \mathbb{E} \tilde{g}(\w_k) \leq \epsilon \mathbb{E}(\u_k - \nabla F(\w_k)) \leq \epsilon \sqrt{d} \sigma_k$. 
Hence, 
\begin{align*}
\mathbb{E} \norm{ \tilde{g}(\w_{k+1}) - \nabla F(\w_{k+1})} \leq (3 + 2 \gamma + \epsilon) \sqrt{d} \sigma_k.
\end{align*}
Thus, if $\norm{ \nabla F(\w_{k+1})} > \norm{ \tilde{g}(\w_{k+1}) - \nabla F(\w_{k+1})}$, we have $$\sin \alpha = \sup_{\forall k} \Big\{ \frac{(3 + 2 \gamma + \epsilon) \sqrt{d} \sigma_k}{\norm{\nabla F(\w_k)}} \Big\}.$$
This completes the proof of verifying Condition i).

\smallskip
Next, we verify Condition ii), i.e., the $r$-th moment of $\tilde{g}(\w)$ is bounded by a linear growth with respect to $\norm{\w}^r$, $r=2,3,4$. 
Note that this is exactly what has been proved in Lemma~\ref{lem_rth_moment_aggr}, which implies that Condition ii) is satisfied.
This completes the proof.
\vspace{-.255in}
\begin{flushright}
\QEDopen
\end{flushright}


{\em Proof of Theorem~\ref{thm_as_converge}.}
The proof of Theorem~\ref{thm_as_converge} follows a similar framework to that used in \cite{bottou1998online} to prove stochastic approximation without attacks. 
We show that, using our aggregation rule in Algorithm~1, the same convergence result can still be obtained under Byzantine attacks.
Let $\mathcal{P}_k$ denote the $\sigma$--algebra (i.e., the information available) available up to iteration $k$. 
Applying the first-order Taylor expansion on $F(\cdot)$ and using the update step (\ref{eqn_weight_update}), we have:
\begin{align*}
&\!F(\w_{k+1}) \!-\!\! F(\w_k) \!\leq\!\! -2 \eta_k \langle \tilde{g}(\w_k), \!\nabla F(\w_k) \rangle \!+\! \eta_k^2 \tilde{g}^2(\w_k) K_1, 
\end{align*}
where $K_1$ is a constant that upper bounds the second derivative (cf. Assumption~\ref{assum5}).
Then, we have
\begin{multline} \label{eqn_one_step_diff}
\mathbb{E} [F(\w_{k+1}) - F(\w_k) | \mathcal{P}_k] \stackrel{(a)}{\leq} -2 \eta_k (1 - \sin \alpha)(\nabla F(\w_k))^2 \\
+ \eta_k^2 \mathbb{E}[\tilde{g}^2(\w_k)] K_1 \stackrel{(b)}{\leq} \eta_k^2 K_2 K_1,
\end{multline}
where $(a)$ is due to $\tilde{g}(\w_k)$ is $(\alpha, q)-$Byzantine resilient and $(b)$ follows from the global confinement of $\w_k$.
Let $\delta_k(x) =1$ if $x>0$ or $0$ otherwise.
Then, taking expectation on both sides of (\ref{eqn_one_step_diff}) yields $\mathbb{E} [\delta_{k}(F(\w_{k+1}) - F(\w_k))] \leq \eta_k^2 K_2 K_1$.
Note that the right-hand side is a convergent  infinite sum due to step-sizes Assumption~\ref{assum4}. 
By quasi-martingale convergence theorem \cite{bottou1998online}, the sequence $F(\w_k)$ converges almost surely, i.e. 
$F(\w_k) \xrightarrow[]{a.s.} F_{\infty}$.
Summing (\ref{eqn_one_step_diff}) over $k = 1, 2, ..., \infty$, the convergence of $F(\w_k)$ implies that:
\begin{align} \label{eqn_eta_grad2}
  \sum_{k=1}^{\infty} \eta_k (\nabla F(\w_k))^2 < \infty \quad a.s. 
\end{align}
Now define $\hat{g}_k = \norm{\nabla F(\w_k)}^2$, following the same process, we can obtain:
$\mathbb{E} [\delta_k (\hat{g}_{k+1} - \hat{g}_k)] = \mathbb{E} [\delta_k \mathbb{E} [\hat{g}(\w_{k+1}) - \hat{g}(\w_k) | \mathcal{P}_k]] 
\leq \eta_k (\nabla F(\w_k))^2 + \eta_k^2 K_2 K_3.$
The two terms on right-hand side are summands of convergent infinite sums by Assumption~\ref{assum4}. 
Again, by quasi-martingale convergence theorem, we have
$\hat{g}_k \xrightarrow[]{a.s.} 0, \nabla F(\w_k) \xrightarrow[]{a.s.} 0$.
\vspace{-.26in}
\begin{flushright}
\QEDopen
\end{flushright}

{\em Proof of Proposition~\ref{prop_TimeComplexity}.}
To obtain the median of an $m$-sized array, the average time complexity is $O(m)$. 
Repeating this process $d$ times in our aggregation rule yields a time-complexity $O(md)$. 
Plus, the time-complexity of comparing the median and each returned gradient $O(md)$. 
Thus, the total time-complexity is $O(md)$.
\vspace{-.26in}
\begin{flushright}
\QEDopen
\end{flushright}



\end{document}